\pdfoutput=1
\documentclass[11pt]{article}

\usepackage[final]{acl}

\usepackage{times}
\usepackage{latexsym}
\usepackage{tipa}
\usepackage{eso-pic}
\usepackage[T1]{fontenc}
\usepackage[utf8]{inputenc}

\usepackage{microtype}

\usepackage{inconsolata}

\usepackage{graphicx}
\usepackage{tipa}
\usepackage{amsmath}
\usepackage{booktabs}
\title{Grammatical Error Correction for Low-Resource Languages: The Case of Zarma}

\author{
\textbf{Mamadou K. Keita\textsuperscript{1}},
\textbf{Adwoa Bremang\textsuperscript{2}},
\textbf{Huy Le\textsuperscript{1}}
\\
\textbf{Dennis Owusu\textsuperscript{2}},
\textbf{Marcos Zampieri\textsuperscript{3}},
\textbf{Christopher Homan\textsuperscript{1}}
\\
\\
\textsuperscript{1}Rochester Institute of Technology,
\textsuperscript{2}Ashesi University,
\textsuperscript{3}George Mason University
}

\begin{document}
\maketitle

\begin{abstract}
Grammatical error correction (GEC) aims to improve text quality and readability. Previous work on the task  focused primarily on high-resource languages, while low-resource languages lack robust tools. 
To address this shortcoming, we present a study on GEC for Zarma, a language spoken by over five million people in West Africa. We compare three approaches: rule-based methods, machine translation (MT) models, and large language models (LLMs). We evaluated GEC models using a dataset of more than 250,000 examples, including synthetic and human-annotated data. Our results showed that the MT-based approach using M2M100 outperforms others, with a detection rate of 95.82\% and a suggestion accuracy of 78.90\% in automatic evaluations (AE) and an average score of 3.0 out of 5.0 in manual evaluation (ME) from native speakers for grammar and logical corrections. The rule-based method was effective for spelling errors but failed on complex context-level errors. LLMs---Gemma 2b and MT5-small---showed moderate performance. Our work supports use of MT models to enhance GEC in low-resource settings, and we validated these results with Bambara, another West African language.

\end{abstract}

\section{Introduction}

Grammatical error correction (GEC) is important to improving text quality in different scenarios. High-resource languages have well-developed GEC tools \citep{Bryant_2023, ng2025conll2013sharedtaskgrammatical} while low-resource languages have few resources. Zarma---a Nilo-Saharan language spoken by over five million people, primarily in Niger, where it is a major national language, as well as in Nigeria, Mali, and Burkina Faso \citep{lewis2016ethnologue}---faces these challenges.

Zarma lacks a standardized orthography and large annotated datasets. To create language technology for low-resource languages like Zara, developers need approaches that handle minimal resources and non-standard writing \citep{tapo2020neural,tapo2025bayelemabaga}. Large language models (LLMs) hold promise as they are capable of handling various languages and tasks \citep{brown2020language}, but performance on very low-resource languages can be limited. LLMs rely heavily on data, which is scarce for Zarma.

To address these important challenges in low-resource NLP, we investigate GEC approaches for Zarma. We evaluate three approaches: a rule-based method, a machine translation (MT)-based model, and LLMs. We also test with Bambara, another West African language, to confirm generalization. 
This paper addresses three key research questions (RQs):

\begin{description}
\item[RQ1:] \textit{Do LLMs outperform traditional rule-based and MT models in Zarma GEC?}
\item[RQ2:] \textit{What are the strengths and limitations of each approach, given the non-standard orthography and limited data?}
\item[RQ3:] \textit {Does the best-performing approach replicate for another low-resource language---Bambara?}
\end{description}

The main contributions of this paper are as follows:
\begin{enumerate}
\item A comparison of three GEC methods for Zarma. The MT-based approach---M2M100---provided the best results, with a detection rate of 95.82\% and a suggestion accuracy of 78.90\% in automatic evaluations (AE) and an average score of 3.0 out of 5.0 in manual evaluation (ME).
\item A dataset of 250,000 synthetic and human-annotated Zarma examples~\footnote{The latest version of the dataset can be found here: \url{27Group/noisy_zarma}}.
\item Experiments with Bambara to confirm replicability.

\end{enumerate}

\section{Related Work}
%GEC is a core language technology. It continues to be a important problem for low-resource languages to solve. In addition to a lack of data, many low-resource languages have only developed writing systems in the past 100 years or so, and consequently grammar is often not widely standardized. 

\citet{cisse-sadat-2023-automatic} present an approach for spellchecking Wolof, a language primarily spoken in Senegal. Their algorithm combines a dictionary lookup with an edit distance metric Levenshtein distance algorithm \citep{levenshtein1966binary} to identify and correct spelling errors.

\citet{dabasoftware} developed Daba, a software package for grammar and spellchecking in Manding languages. This work employs a rule-based system focusing on morphological analysis, addressing the agglutinative nature of Manding languages. 

Researchers have explored the use of LLMs for language-specific tasks, including GEC, demonstrating their adaptability beyond high-resource languages. For instance, a study by \citet{gomez-etal-2023-low} showed that the MT5 model \citep{xue-etal-2021-mt5}, a multilingual transformer model pre-trained on a massive dataset of text and code, could be effectively fine-tuned for GEC in Ukrainian. \citet{song2023gee} present an innovative application of LLMs---specifically GPT-4 \citep{openaiChatGPT4}---for generating explanations for grammatical errors. Another promising approach to GEC leverages pre-trained multilingual MT models. The study by \citet{luhtaru-etal-2024-error} introduced the \emph{No Error Left Behind} approach, which uses models like MT5 and NLLB \citep{nllbteam2022language}. The fine-tuning process involves adapting the pre-trained MT model to treat error correction as a \emph{\textbf{translation}} task, where the source language is the incorrect sentence and the target language is the corrected sentence. The complexity of balancing precision and recall on synthetic data remains an open problem.

\section{Methods}

\subsection{GEC with LLMs}

LLMs have significantly advanced NLP, showing capabilities in multitasking, single-shot, and multilingual learning. These models, extensively pre-trained on diverse datasets, demonstrate a remarkable ability to grasp nuanced aspects of language, reasoning, and context \citep{brown2020language, raffel2020exploring}. This section outlines our methodology for using LLMs to develop a GEC tool for Zarma. The proposed GEC tool is designed to function independently of predefined grammar rules or lexicons, using the models' few-shot learning capabilities for enhanced efficiency in low-resource scenarios.

\subsubsection{Implementation}
We used two distinct approaches for LLM-based GEC:
\subsubsection*{Instruction and Error Explanation Fine-tuning}
This method involves embedding training data within a contextual sentence structure.  We embed examples in a prompt format:
\[
\begin{array}{l}
\label{llm prompt structure}
\text{\textbf{Prompt}: ``Zarma sentence: [\textit{Incorrect Sentence}],} \\
\text{Correct the zarma sentence: [\textit{Correct Sentence}]} \\
\text{Error Causes: [\textit{Error Cause}].''}
\end{array}
\]

This format helps the model reason about corrections and reflect on why errors occur, as demonstrated in previous research \citep{schick2021exploiting, wei2021finetuned}.

\subsubsection*{Non-Prompt Fine-tuning Using Aligned Sentences} 
We also fine-tuned LLMs on parallel incorrect and correct sentences. This leverages the model's ability to align corrupted text with corrected forms.

\subsection{GEC with MT Models}
Using MT models for GEC represents a promising avenue for addressing GEC challenges in low-resource languages such as Zarma. MT models---particularly those pre-trained on multilingual datasets---offer strong capabilities for processing text across diverse linguistic frameworks.

\subsubsection{Implementation}
We chose the M2M100 model \citep{fan2020englishcentric} for its demonstrated ability to translate between many languages, indicating its potential to capture cross-linguistic patterns relevant to GEC.
We adapt M2M100 for GEC by training it to translate corrupted Zarma text into corrected versions. This approach uses a synthetic corpus generated with a noise script (Section \ref{synthetic data}).
The script introduces various errors, ensuring that the training data effectively represent realistic challenges in real-world Zarma text. This corrupted corpus, paired with the original correct sentences, serves as the training data for M2M100, allowing the model to learn the mapping from incorrect to correct Zarma.
Detailed training settings and evaluation metrics for this MT-based GEC approach can be found in Tables \ref{training-details}, \ref{Manual Evaluation Results}, and \ref{combined-metrics}.

\subsection{Rule-based GEC for Zarma}
\label{gen_inst}
\begin{figure}[ht]
\centering
\includegraphics[width=7.5cm, height=6.4cm]{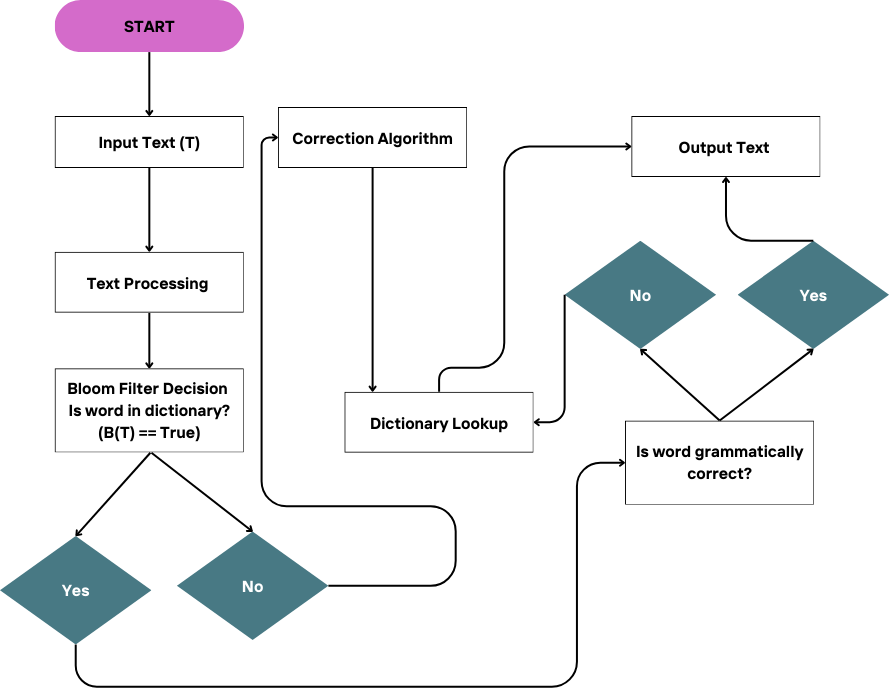}
\caption{Rule-Based GEC tool Workflow}
\label{workflow}
\end{figure}

As a baseline, we built a rule-based system using Levenshtein distance and a Bloom filter \citep{10.1145/362686.362692}, for Zarma (Figure \ref{workflow}). Additionally, we implemented a tool and API in Python. The tool---the first of its kind for Zarma, to our knowledge---is designed for a wide range of users. It offers command-line and graphical user interfaces (GUI) and is much less computationally intensive than the LLM-based approach. Moreover, our results show that it provides a competitive spell correction performance compared to the LLM- and MT-based approaches. To ensure the accessibility of the tool, we plan a public release upon acceptance.

The GEC process for Zarma text follows a structured sequence of steps, starting with the input text \( T \), passing through a Bloom filter decision \( B \) to determine if the word exists in the dictionary, leading to either a grammatical check or a correction algorithm, followed by a final dictionary lookup and correction \( C \), as needed:

\begin{equation}
C(T) = \begin{cases}
T & \text{if } D(B(T)) = \text{True}, \\
\text{Correct}(T) & \text{if } D(B(T)) = \text{False}.
\end{cases}
\end{equation}

The detailed process includes:

\paragraph{Text Processing}
The system begins by processing the text, where the Zarma content is segmented into words, punctuation, and spaces using regular expressions. This initial step ensures that every part of the text is ready for individual examination---a necessity given Zarma's linguistic intricacies.

\paragraph{Bloom Filter}
The Bloom filter---known for its space and time efficiency---performs probabilistic checks on whether a word might be in the dictionary. For a given word \( w \), it uses a set of hash functions to probe various positions in a bit array. If all checked positions are flagged, \( w \) might be in the dictionary:

\begin{equation}
B(w) = \bigwedge_{i=1}^{k} \text{bit}[h_i(w)],
\end{equation}

If \( B(T) = \text{False} \), the system assumes the word is not in the dictionary and immediately moves to the correction algorithm to suggest a correction. This process ensures efficiency when dealing with unknown words.

\paragraph{Dictionary Lookup}
At the core of the system is the trie-based dictionary lookup. We built this dictionary using the Feriji dataset dictionary \citep{keita-etal-2024-feriji}. 
If the Bloom filter decision is positive (\( B(T) = \text{True} \)), the system proceeds to a Grammatical Check to verify if the word is grammatically correct. If the word is found to be grammatically incorrect, or if \( B(T) = \text{False} \) (indicating the word is not in the dictionary), the system performs a Dictionary Lookup to confirm or correct the word.

\paragraph{Grammar Rules}
The system incorporates a set of Zarma grammar rules, a task made challenging by the lack of a standard writing format. Zarma is classified as an agglutinative language, meaning it depends heavily on attaching suffixes to root words to indicate grammatical features like plurality and definiteness (e.g., the definite plural suffix \textbf{-ey}). This morphology imposes specific constraints on our rule-based approach: the rules must be robust enough to parse and validate these suffix chains rather than treating words as static units. This phase scrutinizes elements like consonant rules and vowel lengths, essential for keeping the text accurate to Zarma's grammatical essence despite its absence of a writing standard.

\paragraph{Correction Algorithm}
When an error is detected, either because the Bloom filter returned false or the grammatical check identified an issue, the Correction Algorithm is activated. This component leverages the Levenshtein distance \( L \) to identify the smallest number of edits---insertions, deletions, or substitutions---required to rectify the erroneous word. For illustration, given a corrupted sentence ``A sindq biri,'' the Levenshtein algorithm's operation can be represented as:

\begin{equation}
\begin{array}{c}
\text{\textbf{Corrupted Sentence}: ``A sindq biri''} \\
\downarrow \\
\text{\textbf{LevSuggest}}(``sindq'') \\
\downarrow \\
\{\text{``sind''}, \text{``sinda''}\}
\end{array}
\end{equation}

Here, ``sindq'' undergoes comparison with the lexicon, and ``sind'' and ``sinda'' are suggested based on their proximity in terms of minimal edit distance. In this context, \( L(a, b) \) calculates the distance between the incorrect string \( a \) and a potential correct string \( b \), considering their lengths \( i \) and \( j \), with \( 1_{(a_i \neq b_j)} \) serving as an indicator function to highlight discrepancies. This mechanism ensures the algorithm's efficiency in providing appropriate corrections, thus enhancing the GEC tool's overall accuracy for Zarma text.

\section{Data}
\label{synthetic data}

To address the scarcity of annotated data in Zarma, we created a combined dataset of 250,000 examples by merging two distinct datasets---a synthetic dataset and a human-annotated ``gold'' dataset. The synthetic dataset contains 248,000 sentence pairs; the human-annotated dataset includes 2,000 sentence pairs. We split this combined dataset into 80\% training, 10\% validation, and 10\% testing sets.

\subsection{Synthetic Data}
\begin{table*}[ht]
\centering
\small
\begin{tabular}{ l p{7cm} }
\hline
\textbf{Correct Sentence} & \textbf{Corrupted Sentences} \\
\hline
\textit{Sintina gaa Irikoy na beena da ganda taka.} & Sintina gaa Irikog na beena da ganda taka. \\ 
& Sintina gaa Irikoy na been da ganda taka. \\ 
& Sintina aga Irikoy na beena da ganda taka. \\ 
& Sintina gaa Irikoy na beena ea ganda taka. \\
\hline
\end{tabular}
\caption{Snapshot of The Synthetic Data}
\label{zarma-dataset}
\end{table*}

We applied a noise script to the Feriji dataset \citep{keita-etal-2024-feriji} to generate synthetic errors. This script introduces typographical and grammatical errors into Zarma sentences---errors that mimic human mistakes. To ensure these errors reflect real-world patterns, we worked with five native Zarma speakers---the same individuals who later conducted the manual evaluation (ME) as described in Section~\ref{evaluators}. We provided them with audio recordings of 46 Zarma sentences from everyday conversations, which they transcribed into written Zarma---with instructions to not attempt to correct any mistake. All typed in French at an average of 38 words per minute using an AZERTY keyboard. We compared their transcriptions with the correct Zarma text and analyzed the frequency of error types. The study revealed common errors: insertions (e.g., ``Irikoy'' to ``\textbf{Irikoyu}''), and deletion (e.g., ``alfarey'' to ``\textbf{alfary}''); with frequencies of 23\% and 12\%, respectively. Minor transpositions and double vowels were noticed too---less than 2\% and 7\% respectively. The AZERTY keyboard layout and typing speed influence error types such as insertions due to adjacent key proximity---e.g, `u' near `y' in `Irikoyu'). Also, the average error per sentence was 1 error for short sentences (1-5 words), 3 errors for long sentences (5+ words).

The noise function $\mathbf{N}$ consists of four operations: deletions $\mathbf{\delta}$, insertions $\mathbf{\mu}$, substitutions $\mathbf{\sigma}$, and transpositions $\mathbf{\tau}$. For example, applying a substitution operation to the original sentence $\mathbf{SZ} = \textit{A go koy fuo}$ (``He is going home'') might yield $\mathbf{CSZ} = \textit{A ga koy fuo}$ (``He will go home''), introducing a grammatical error by changing the tense.

\begin{equation}
  \begin{aligned}
    &\mathbf{SZ} = ``A \, go \, koy \, fuo'' \\
    &\mathbf{N}(\mathbf{SZ}) = \mathbf{SZ} \xrightarrow{\sigma(``go'' \to ``ga'')} ``A \, ga \, koy \, fuo'' \\
    &\mathbf{CSZ} = ``A \, ga \, koy \, fuo'' \\
    &\mathbf{SZ} \xrightarrow{N} \mathbf{Correct} \, \mathbf{CSZ}
  \end{aligned}
\end{equation}

Furthermore, to enrich the dataset with diverse linguistic nuances, the script generates four corrupted variants for each correct sentence in the dataset. A snapshot of the synthetic data appears in Table~\ref{zarma-dataset}.

\subsection{Human-Annotated Data}
The human-annotated dataset was curated by native Zarma speakers who introduced logical and grammatical (LG) errors into Zarma sentences and provided justifications for each alteration. This dataset exposes the models to a broader array of error types---particularly logical errors---and their corresponding corrections, enhancing the models' capacity to generalize and accurately correct unseen texts in Zarma. An example from the human-annotated dataset is shown in Table~\ref{zarma-dataset-modification}

\begin{table}[ht]
\small
\centering
\begin{tabular}{ p{1.5cm} p{1.5cm} p{3cm} }
\hline
\textbf{Incorrect Sentence} & \textbf{Correction} & \textbf{Error Explanation} \\
\hline
\textit{Souba, Ay koy Niamey} & Souba, Ay ga koy Niamey & ``\textbf{Souba}'' means tomorrow, and therefore the tense must be in the future using the future tense marker ``\textbf{ga}'' after the subject ``\textbf{Ay}.'' \\
... & ... & ... \\
\hline
\end{tabular}
\caption{Snapshot of the human-annotated dataset}
\label{zarma-dataset-modification}
\end{table}

\section{Experiment}

We conducted a series of experiments to evaluate the effectiveness of different GEC methods for Zarma. We selected three models for training based on their demonstrated proficiency in multilingual tasks and aptitude for few-shot learning: \textbf{Gemma}~\citep{team2024gemma}, \textbf{MT5-small}, and \textbf{M2M100}.
The training was conducted on Google Cloud, using a Deep Learning Virtual Machine equipped with NVIDIA GPUs T4 x2. Due to computational resource constraints, we applied QLoRA quantization \citep{dettmers2023qlora} to the Gemma model to reduce memory requirements. For MT5-small and M2M100, we used smaller variants of the models to ensure feasible training times.
For LLM-based methods (Gemma and MT5-small), we structured the combined dataset as described in Section \ref{synthetic data}, to incorporate both incorrect sentences and their corrections along with error explanations. For the MT-based method (M2M100), we adopted a MT task-specific format, using aligned pairs of incorrect and correct Zarma sentences without explicit error explanations.
The detailed training settings for each model are presented in Table \ref{training-details}.

\section{Results \& Comparative Analysis}

To assess the effectiveness of each GEC method for the types of errors addressed in this paper (as detailed in Section~\ref{error_types}), we conducted an evaluation consisting of two parts: an automatic evaluation (AE) and ME by native Zarma speakers. We included the ME because AE tools are primarily developed and tested on high-resource languages. Therefore, their scores may not reliably reflect performance in low-resource settings such as Zarma. The ME allows us to obtain assurances from language speakers regarding the actual effectiveness of the methods.

\subsection{Automatic Evaluation Results}

In the AE, we employed two sets of metrics to evaluate our techniques. The first set consisted of common machine learning metrics, including the GLEU score \citep{napoles-etal-2015-ground} and the M\textsuperscript{2} score \citep{dahlmeier-ng-2012-better}. The second set focused on spelling correction metrics---measuring the detection rate, suggestion accuracy, and false positive rate. The results are presented in Table~\ref{AE Results Tables}.

\begin{table*}[ht]
\centering
\small
\begin{tabular}{p{1.5cm}p{2cm}p{2cm}p{2cm}p{2cm}p{2cm}}
\toprule
\textbf{Methods} & \multicolumn{5}{c}{\textbf{AE Evaluation}} \\
\cmidrule(r){2-6}
& \multicolumn{2}{c}{\textbf{Common ML Metrics}} & \multicolumn{3}{c}{\textbf{Spelling Correction}} \\
\cmidrule(r){2-3} \cmidrule(r){4-6}
& \textbf{GLEU} & \textbf{M²} & \textbf{Detection} & \textbf{Correction} & \textbf{False Positive} \\ 
\midrule
Rule-based & 0.3124 & 0.401 & 100\% & 96.27\% & 2.5\% \\
Gemma 2b & 0.4572 & 0.528 & 76.19\% & 43.28\% & 15.3\% \\
MT5-small & 0.6203 & 0.731 & 90.62\% & 57.15\% & 8.7\% \\
M2M100 & 0.7896 & 0.914 & 95.82\% & 78.90\% & 4.2\% \\ 
\bottomrule
\end{tabular}
\caption{AE metrics scores}
\label{AE Results Tables}
\end{table*}

\subsubsection{Common Machine Learning Metrics}
The GLEU and M\textsuperscript{2} scores provide an overall assessment of the models' performance in correcting grammatical errors by comparing the corrected sentences to reference sentences. The rule-based method achieved modest scores (GLEU: 0.3124, M\textsuperscript{2}: 0.401), reflecting its effectiveness in spelling correction but limitations in handling complex grammatical errors. The higher scores achieved by the M2M100 model (GLEU: 0.7896, M\textsuperscript{2}: 0.914) indicate its superior performance among the tested models. The LLM-based models---Gemma 2b and MT5-small---obtained moderate scores, which shows their ability to correct some errors but not as effectively as the M2M100 model.

\subsubsection{Spelling Correction Metrics}
For the spelling correction evaluation, we used a set of 3,539 sentences from the Feriji Dataset, introducing typographical errors using our corruption script (as described in Section \ref{synthetic data}). We measured three key metrics:

\begin{itemize} 
\item \textbf{Detection Rate}: The percentage of errors that the method successfully identified and attempted to correct. 
\item \textbf{Suggestion Accuracy}: The percentage of corrections suggested by the method that were accurate. 
\item \textbf{False Positive Rate\footnote{While we did not evaluate the models on entirely correct sentences, our False Positive Rate reflects over-corrections within mixed correct-and-incorrect datasets, approximating real-world GEC scenarios where fully pristine text is rare.}}: The percentage of correct words incorrectly flagged as errors. Here, a false positive is counted not only when an entire sentence is flagged, but also when parts of a sentence that are correct are unnecessarily altered. In other words, if a sentence contains both correct and incorrect segments, any modification of the correct segments is considered a false positive.

\end{itemize}

As shown in Table \ref{AE Results Tables}, the rule-based method achieved a detection rate of 100\% and a suggestion accuracy of 96.27\%, with a low false positive rate of 2.5\%. This high performance is primarily due to its dictionary---built from the Feriji Dataset---and predefined rules, which allow it to effectively flag and correct errors within its scope while rarely misidentifying correct words. The M2M100 model also performed well, with a detection rate of 95.82\%, a suggestion accuracy of 78.90\%, and a false positive rate of 4.2\%, indicating its capability to identify and correct spelling errors in a with minimal over-correction. The LLM-based models---Gemma 2b and MT5-small---showed lower detection rates and suggestion accuracies, coupled with higher false positive rates. Gemma 2b detected 76.19\% of errors with a suggestion accuracy of 43.28\% and a false positive rate of 15.3\%, while MT5-small detected 90.62\% of errors with a suggestion accuracy of 57.15\% and a false positive rate of 8.7\%. These results suggest that while LLMs can correct some spelling errors, they are less effective than the MT-based and rule-based methods in this task and are more prone to incorrectly flagging correct words. The reasons of this poor performance of LLMs is due to the fact they were not pre-trained on Zarma, thereby limiting their accuracies in Zarma related tasks.

\subsection{Performance on Synthetic and Human-Annotated Subsets}

To further explore how each method handles different data sources, we tested performance on the synthetic and human-annotated subsets described in Section~\ref{synthetic data}. Table~\ref{subset-performance} presents the results side by side.

\begin{table*}[ht]
\centering
\small
\begin{tabular}{p{1.5cm}p{2cm}p{2cm}p{2cm}p{2cm}}
\toprule
\textbf{Methods} & \multicolumn{2}{c}{\textbf{Synthetic Subset}} & \multicolumn{2}{c}{\textbf{Human-Annotated Subset}} \\
\cmidrule(r){2-3} \cmidrule(r){4-5}
& \textbf{GLEU} & \textbf{M²} & \textbf{GLEU} & \textbf{M²} \\
\midrule
Rule-based & 0.3150 & 0.405 & 0.2900 & 0.375 \\
Gemma 2b & 0.4601 & 0.531 & 0.4213 & 0.492 \\
MT5-small & 0.6257 & 0.738 & 0.5812 & 0.692 \\
M2M100 & 0.7923 & 0.917 & 0.7618 & 0.891 \\
\bottomrule
\end{tabular}
\caption{Performance metrics on synthetic and human-annotated subsets}
\label{subset-performance}
\end{table*}

Table~\ref{subset-performance} reveals a pattern---all models, including the rule-based method, score lower on the human-annotated subset than on the synthetic subset. For the rule-based method, GLEU drops from 0.31 to 0.2900 and M² from 0.40 to 0.375, indicating its reduced capability with complex logical errors present in human-annotated data. This difference arises because human-annotated data includes complex logical and context-level errors---challenges beyond the simpler typographical errors in the synthetic set. Nevertheless, M2M100 maintains its lead across both subsets. Its GLEU drops from 0.7923 to 0.7618 and M² from 0.917 to 0.891, yet these scores still outperform Gemma 2b and MT5-small by 5\%--10\%. This consistency highlights M2M100’s strength in tackling diverse error types.

\subsection{Manual Evaluation Results}

Given the limitations of AE metrics in low-resource settings, we conducted the ME involving five native Zarma speakers. The evaluators assessed how well each method corrected LG errors, using a scale from 1 to 5, where 1 indicates a poor correction and 5 indicates an excellent correction. The evaluation was performed on a set of 300 sentences from the Feriji Dataset, which were manually corrupted by our annotators to include LG errors. Detailed information about the evaluation protocol and the recruitment of evaluators can be found in Section \ref{evaluators}. The results are presented in Table \ref{Manual Evaluation Results}.

\begin{table}[ht]
\centering
\small
\begin{tabular}{p{2cm}p{2cm}p{2cm}}
\toprule
\textbf{Methods} & \multicolumn{2}{c}{\textbf{ME avg (1-5)}} \\
\cmidrule(r){2-3} 
& \textbf{Logical Errors Correction} & \textbf{Sentence Improvement} \\
\midrule
Rule-based & 0.4 & 0 \\
Gemma 2b & 1 & 0 \\
MT5-small & 1.7 & 1 \\
M2M100 & 3 & 2.5 \\
\bottomrule
\end{tabular}
\caption{Manual Evaluation}
\label{Manual Evaluation Results}
\end{table}

The rule-based method struggled with LG errors, achieving low scores (0.4 for logical error correction and 0.0 for sentence improvement). This outcome is expected, as the rule-based approach relies on predefined rules and lacks the ability to understand the context or semantics of sentences. The M2M100 model outperformed the other methods, achieving higher scores in both logical error correction (3.0) and sentence improvement (2.5). This demonstrates that the MT-based approach is more effective at handling LG errors and improving the overall quality of the sentences. The LLM-based models achieved moderate scores, with MT5-small performing better than Gemma 2b. MT5-small received scores of 1.7 for logical error correction and 1.0 for sentence improvement, suggesting some capability in handling LG errors, but not as effectively as the M2M100 model. We observed that the models faced varying levels of difficulty with different types of LG errors. For instance, they were generally better at correcting verb tense errors than subject-verb agreement errors.

\subsection{Zero-Shot Scores}
\begin{table*}[ht]
\centering
\small
\begin{tabular}{p{1.5cm}p{1cm}p{1cm}p{1.3cm}p{1.5cm}p{2cm}p{1cm}p{1.5cm}}
\toprule
\textbf{Methods} & \multicolumn{5}{c}{\textbf{AE Evaluation}} & \multicolumn{2}{c}{\textbf{Manual Evaluation avg (1-5)}} \\
\cmidrule(r){2-6} \cmidrule(r){7-8}
& \multicolumn{2}{c}{\textbf{Common ML Metrics}} & \multicolumn{3}{c}{\textbf{Spelling Correction}} & \textbf{Logical Errors Correction} & \textbf{Sentence Improvement} \\
\cmidrule(r){2-3} \cmidrule(r){4-6} \cmidrule(r){7-8}
& \textbf{GLEU} & \textbf{M²} & \textbf{Detection} & \textbf{Correction} & \textbf{False Positive} & & \\
\midrule
Rule-based & 0.2800 & 0.365 & 94.17\% & 81.04\% & 3.8\% & 0 & 0 \\
Gemma 2b & 0.3926 & 0.481 & 53.93\% & 26.52\% & 18.2\% & 0.6 & 0 \\
MT5-small & 0.5018 & 0.657 & 73.64\% & 58.31\% & 10.5\% & 0.8 & 1 \\
M2M100 & 0.7683 & 0.904 & 92.27\% & 78.34\% & 5.1\% & 2.7 & 1.3 \\
\bottomrule
\end{tabular}
\caption{Performance Evaluation on Zero-Shot Dataset}
\label{combined-metrics}
\end{table*}

\begin{table*}[h]
\centering
\small
\begin{tabular}{p{1.5cm}p{2cm}p{2cm}p{2cm}p{2cm}}
\toprule
\textbf{Methods} & \multicolumn{4}{c}{\textbf{AE Evaluation}} \\
\cmidrule(r){2-5}
& \multicolumn{2}{c}{\textbf{Common ML Metrics}} & \multicolumn{2}{c}{\textbf{Spelling Correction}} \\
\cmidrule(r){2-3} \cmidrule(r){4-5}
& \textbf{GLEU} & \textbf{M²} & \textbf{Detection} & \textbf{Correction} \\ 
\midrule
MT5-small & 0.5713 & 0.643 & 87.45\% & 52.91\% \\
M2M100 & 0.7293 & 0.851 & 94.64\% & 68.18\% \\ 
\bottomrule
\end{tabular}
\caption{Performance Metrics for Bambara}
\label{Experiment Results for Bambara}
\end{table*}

To further assess the generalization capabilities of our models, we conducted a zero-shot evaluation using a separate dataset. For that, we gathered a separate set---150 sentences---of Zarma texts from books that had not been used during training. We then applied the same corruption script and asked the annotators to include some LG errors, containing sentences with words and grammatical structures not present in the training data. This evaluation aimed to understand how well the models can handle new or unseen data, which is a common scenario in low-resource language settings. The results of the zero-shot evaluation are presented in Table \ref{combined-metrics}.

As expected, the rule-based method showed a decrease in performance compared to the controlled evaluation. Its GLEU score dropped to 0.2800 and M² to 0.365, with a detection rate of 94.17\%, suggestion accuracy of 81.04\%, and a false positive rate of 3.8\%. This decline is due to the method's reliance on a predefined dictionary and rules, which may not cover new words or grammatical constructions present in the zero-shot dataset, though it maintains a low rate of incorrect flagging. Additionally, the ME scored it with 0 points for both logical errors correction and sentence improvement.

The M2M100 model again demonstrated higher performance among the tested models in the zero-shot setting. It achieved a GLEU score of 0.7683 and an M² score of 0.904, indicating strong overall correction capability. Its detection rate was 92.27\% with a suggestion accuracy of 78.34\% and a false positive rate of 5.1\%, showing that it can effectively identify and correct errors in unseen data while keeping over-correction low. Furthermore, in ME, it received 2.7 for logical errors correction and 1.3 for sentence improvement.

The LLM-based models, showed reduced performance in the zero-shot evaluation. Gemma 2b's detection rate was 53.93\% with a suggestion accuracy of 26.52\% and a false positive rate of 18.2\%, while MT5-small achieved a detection rate of 73.64\%, a suggestion accuracy of 58.31\%, and a false positive rate of 10.5\%. In terms of ME, Gemma 2b scored 0.6 for logical errors correction and 0 for sentence improvement, while MT5-small achieved 0.8 and 1, respectively. These higher false positive rates indicate a greater tendency to misidentify correct words as errors in unseen data.

Our results suggest that MT models trained on multilingual datasets hold significant potential for improving GEC in low-resource languages. The ME scores further validate that the M2M100 model excels not only in automated metrics but also in human judgments, particularly in complex correction tasks. The lower accuracy of LLMs likely arises from their \textbf{\textit{lack of initial pre-training on Zarma and the use of quantization during training}}, which may impact performance due to limited computational resources in our low-resource setting. In contrast, M2M100 performed well despite these constraints. \textbf{We hypothesize} that in higher computational resource settings, \textbf{LLMs could outperform} all other methods, leveraging their capacity for nuanced language understanding when adequately pre-trained and resourced. 

Nonetheless, to validate the reproducibility and robustness of our methods, we conducted additional experiments with Bambara---a language from a different linguistic family. The results of this experiment are detailed in Section~\ref{more experiments}.

\section{Further Experiment with other Languages}
\label{more experiments}

After obtaining promising results for Zarma using LLM and MT-based approaches, we conducted further experiments to validate the reproducibility of these methods---using the M2M100 and MT5 models. We selected the Bambara language for this experiment because it belongs to a different linguistic family, which allows us to evaluate the performance of the approaches on a language outside the Nilo-Saharan family. We used the Bayelemabaga dataset \citep{bayelemabagamldataset2022}---split into 80\% 10\% 10\%---for this experiment. The same data preparation process described in the methodology section was followed; however, we excluded any human-annotated data to focus only on spelling correction performance. The results are presented in Table \ref{Experiment Results for Bambara}.

The Bambara experiment demonstrated that the MT-based approach outperformed the LLMs-based one regarding word-level correction metrics. The MT-based approach achieved a detection rate of 94.64\% and a suggestion accuracy of 68.18\%. In contrast, the LLMs-based approach detected 87.45\% of errors and suggested corrections with 52.91\% accuracy. The results from the Bambara experiment highlight the potential of both LLMs and MT models to improve GEC for low-resource languages significantly. However, they also emphasize the necessity for continued expanding and diversifying training datasets and pre-training for LLMs.

\section{Conclusion and Future Work}
We presented a comparison of rule-based, LLM-based, and MT-based methods for Zarma GEC. Our results provide answers to our research questions. Regarding RQ1, we found that MT models (M2M100) outperform both rule-based systems and LLMs like Gemma and MT5 in Zarma GEC. For RQ2, we observed that rule-based methods excel at simple spelling correction but fail at context-level errors, whereas MT models handle both effectively. Finally, for RQ3, our experiments with Bambara confirm that the MT-based approach is replicable and effective for other low-resource West African languages. This research establishes a foundation for future GEC development in other low-resource languages. The approaches covered in this study are starting point references and can be further improved using sophisticated methods.

Future work includes hybrid models that combine rule-based precision with MT and LLM adaptability, data augmentation to improve performance, continuous learning mechanisms, resource optimization, and cross-lingual transfer learning to support a wider range of languages.

%\newpage

\begin{table*}[ht!]
  \centering
  \tiny
  \begin{tabular}{p{1.3cm}p{1.5cm}p{1.5cm}p{1cm}p{1.5cm}p{1cm}p{1.5cm}}
    \toprule
    \textbf{Models} & \textbf{Parameters} & \multicolumn{5}{c}{\textbf{Training Details}} \\
    \cmidrule(r){3-7}
     &  & \textbf{QLoRA} & \textbf{Epochs} & \textbf{Batch Size} & \textbf{GPU Used} & \textbf{Lr} \\
    \midrule
    Gemma 2b & 2 billion & Applied & 2 & 8 & P100 & $2 \times 10^{-4}$ \\
    MT5-small & 300 million & Not Applied & 10 & 16 & T4x2 & $2 \times 10^{-5}$ \\
    M2M100 & 418 million & Not Applied & 2 & 32 & P100 & $2 \times 10^{-5}$ \\
    \bottomrule
  \end{tabular}
  \caption{Training settings for the models, including epochs, batch size, and optimizer details.}
  \label{training-details}
\end{table*}

\section*{Limitations}

\paragraph{Rule-Based Method}
Rule-based approaches rely on predefined linguistic rules and dictionaries, which limit their adaptability. They do not handle new or unseen words well. In zero-shot scenarios, where the system encounters words or structures beyond its dictionary, it often fails to correct or detect errors. Additionally, the creation and maintenance of linguistic resources is difficult for low-resource languages, which often lack standardized orthography \citep{scannell2007challenges}.

\paragraph{LLMs Method}
LLMs---Gemma 2b and MT5-small---must rely on diverse training data to capture linguistic nuances. Yet, existing LLMs are often trained on data from high-resource languages, which reduces performance in languages like Zarma. A lack of training samples in these languages causes the model to produce lower-quality suggestions \citep{bender2021dangers}. In addition, fine-tuning LLMs requires significant computational resources, which can be a challenge in low-resource settings.

\paragraph{MT Method}
MT models---M2M100 in our studies---learn from parallel corpora that map incorrect text to correct text. In low-resource languages, such parallel data is often lacking or incomplete \citep{tiedemann2020tatoeba}. Even though multilingual models can leverage patterns from other languages, the absence of sufficient in-language data can lead to contextually imprecise corrections. Moreover, these models require targeted fine-tuning and periodic updates as language use evolves.

\paragraph{Synthetic Data Generation}
Our synthetic data generation process---while designed to mimic common human errors based on an initial study with native speakers (as detailed in Section~\ref{synthetic data})---has some limitations. The corruption script may not capture the full nuanced distribution of errors that humans make---particularly complex grammatical or semantic errors that are not easily programmable by rule-based mechanisms. This could lead to models being better prepared for certain types of synthetic errors than for the true wide range of errors encountered in real-world Zarma texts. Consequently, performance on naturally occurring errors might differ from that observed on our synthetic test sets.

\section*{Acknowledgments}

This research was partially supported by the computer science department of Ashesi University, and the Cite de l'Innovation of Niger.

\bibliography{custom}

\appendix
\section{Errors Being Addressed}
\label{error_types}
In this section, we explain the types of errors GEC methods address. We categorize the errors into two main types: word-level correction (spellchecking) and context-level correction. The context-level correction is further divided into logical errors and sentence improvement. Below, we define each error type and provide examples in Zarma with English translations to illustrate them.

\subsection{Word-Level Correction}
Word-Level correction involves identifying and correcting typographical errors in individual words. These errors are usually due to misspellings, incorrect usage of characters, or typographical mistakes. Such errors typically do not alter the grammatical structure of the sentence but affect readability and lexical accuracy.

\begin{itemize}
    \item \textbf{Example:} 
    \begin{itemize}
        \item \textbf{Incorrect Zarma}: \textit{Sintina gaa Irik\textbf{og} na beena da ganda taka.}
        \item \textbf{English Translation (Incorrect)}: In the beginning G\textbf{o}d [misspelled form of God] created the heaven and the earth.
        \item \textbf{Correct Zarma}: \textit{Sintina gaa Irik\textbf{oy} na beena da ganda taka.}
        \item \textbf{English Translation (Correct)}: In the beginning God created the heaven and the earth.
        \item \textbf{Explanation}: The word ``\textit{Irikog}'' is a misspelling of ``\textit{Irikoy}'' (God). Spellchecking corrects this specific lexical error.
    \end{itemize}
\end{itemize}

\subsection{Context-Level Correction}
\label{lg description}
Context-level correction involves errors that go beyond individual words and affect the overall structure, coherence, and meaning of the sentence. These errors require an understanding of the surrounding context to identify and rectify. We categorize these errors into logical errors and sentence improvement.

\subsubsection{Logical Errors}
Logical errors are inconsistencies in sentence meaning often caused by incorrect verb conjugations, subject-verb agreement issues, or the misuse of grammatical markers such as tense or aspect markers. These errors affect the semantic coherence of the sentence and require contextual understanding to correct.

\begin{itemize}
    \item \textbf{Example:} 
    \begin{itemize}
        \item \textbf{Incorrect Zarma}: \textit{Souba, Ay \textbf{koy} Niamey.}
        \item \textbf{English Translation (Incorrect)}: Tomorrow, I \textbf{go} to Niamey.
        \item \textbf{Correct Zarma}: \textit{Souba, Ay \textbf{ga koy} Niamey.}
        \item \textbf{English Translation (Correct)}: Tomorrow, I \textbf{will go} Niamey.
        \item \textbf{Explanation}: The time indicator ``\textit{Souba}'' means ``tomorrow,'' which requires a future tense. The original sentence incorrectly uses ``\textit{koy}'' (go/am going), which is present tense. The correction introduces the future tense marker ``\textit{ga}'' before the verb ``\textit{koy}'' to align the verb tense with the future time reference, resolving the logical inconsistency.
    \end{itemize}
\end{itemize}

\subsubsection{Sentence Improvement}
Sentence improvement focuses on enhancing the overall readability, precision, naturalness, or stylistic quality of a sentence. While the original sentence may already be grammatically correct, improvements aim to make it flow better, be more concise, or adhere more closely to common usage patterns.

\begin{itemize}
    \item \textbf{Example:}
    \begin{itemize}
        \item \textbf{Original Zarma}: \textit{Albarka ma te ni se te\textipa{\ng} ka\textipa{\ng} ni goy.}
        \item \textbf{English Translation}: Blessings be to you for the work that you did. 
        \item \textbf{Improved Zarma}: \textit{Ni goyom albarka.}
        \item \textbf{English Translation}: Blessings on your work / Your work is blessed.
        \item \textbf{Explanation}: The original sentence is grammatically correct but might be perceived as slightly formal or verbose in some contexts. The improved version offers a more concise and idiomatic expression common in Zarma for acknowledging good work.
    \end{itemize}
\end{itemize}

\section{Evaluators Recruitment Process}
\label{evaluators}

To ensure a thorough and culturally appropriate manual evaluation of our GEC methods, we recruited five native Zarma speakers to serve as evaluators, who also assisted in developing the corruption script for synthetic data (Section~\ref{synthetic data}). The primary criterion for selection was their fluency in Zarma---their ability to read and write proficiently in the language. The evaluators were selected through community outreach in Niamey, Niger. All selected evaluators had completed secondary education or higher.

Prior to the evaluation, we conducted a training workshop to familiarize the evaluators with the objectives of the study, the evaluation protocols, and the scoring guidelines. 
During the evaluation process, each evaluator independently assessed a set of 300 sentences that had been manually corrupted to include logical and grammatical errors. They reviewed the corrections made by each GEC method and assigned scores based on accuracy and fluency, where:

\begin{itemize} 
\item \textbf{1} indicates a poor correction with significant errors remaining or introduced. 
\item \textbf{5} indicates an excellent correction with all errors appropriately addressed and the sentence reading naturally. 
\end{itemize}

All evaluators gave informed consent to participate in the study and were compensated for their time and contributions.

\section{Potential Applications}
Our team visited Niamey to present our work to the local Zarma community and gather their feedback. The discussions provided valuable insights into potential applications for our GEC tool and broader language models.

\paragraph{Content Creation and Educational Resources}
One significant comment we received was the potential use of our model to translate coding content and general educational materials for enthusiasts and students. There is a growing interest in technology and programming within the community, but a considerable language barrier exists due to the lack of educational resources in Zarma. By translating and correcting coding tutorials, textbooks, and other educational materials into Zarma, our model can help overcome this challenge.
Additionally, the GEC tool can be used to improve general educational content in Zarma, including textbooks, instructional materials, and online courses across various subjects.

\paragraph{Cultural Preservation and Documentation}
Feedback from the community highlighted the importance of maintaining accurate written records of folklore, oral histories, and traditional knowledge. The GEC tool can support these efforts by providing a means for transcribing and publishing grammatically accurate texts.

\end{document}